\documentclass[AMS,Times1COL]{WileyNJDv5} 
\usepackage{siunitx}
\usepackage{textcomp}

\definecolor{Min1}{rgb}{1, 0.75, 0.7}  
\definecolor{Min2}{rgb}{1, 0.83, 0.7}   
\definecolor{Min3}{rgb}{1, 0.96, 0.7}  

\usepackage{colortbl}

\articletype{Article Type}%

\received{Date Month Year}
\revised{Date Month Year}
\accepted{Date Month Year}
\journal{Journal}
\volume{00}
\copyyear{2023}
\startpage{1}

\begin{document}

\title{UNO: Unified Self‑Supervised Monocular Odometry for Platform‑Agnostic Deployment}

\author[1]{Wentao Zhao}

\author[2]{Yihe Niu}

\author[1]{Yanbo Wang}

\author[1]{Tianchen Deng}

\author[3]{Shenghai Yuan}

\author[4]{Zhenli Wang}

\author[4]{Rui Guo}

\author[1]{Jingchuan Wang}

\authormark{Zhao \textsc{et al.}}
\titlemark{UNO: Unified Self‑Supervised Monocular Odometry for Platform‑Agnostic Deployment}

\address[1]{Institute of Medical Robotics and Department of Automation, Shanghai Jiao Tong University, Shanghai, China.}

\address[2]{School of Mathematical Sciences Shanghai Jiao Tong University, Shanghai Jiao Tong University, Shanghai, China.}

\address[3]{The Centre for Advanced Robotics Technology Innovation (CARTIN), School of Electrical and Electronic Engineering, Nanyang Technological University, Singapore, Singapore}

\address[4]{State Grid Intelligence Technology CO., LTD.}

\corres{Corresponding author Jingchuan Wang, Institute of Medical Robotics and Department of Automation, Shanghai Jiao Tong University, Shanghai, China. \email{jchwang@sjtu.edu.cn};
}

\abstract[Abstract]{This work presents UNO, a unified monocular visual odometry framework that enables robust and adaptable pose estimation across diverse environments, platforms, and motion patterns. Unlike traditional methods that rely on deployment-specific tuning or predefined motion priors, our approach generalizes effectively across a wide range of real-world scenarios, including autonomous vehicles, aerial drones, mobile robots, and handheld devices. To this end, we introduce a Mixture‑of‑Experts strategy for local state estimation, with several specialized decoders that each handle a distinct class of ego-motion patterns. Moreover, we introduce a fully differentiable Gumbel‑Softmax module that constructs a robust inter‑frame correlation graph, selects the optimal expert decoder, and prunes erroneous estimates. These cues are then fed into a unified back‑end that combines pre-trained, scale-independent depth priors with a lightweight bundling adjustment to enforce geometric consistency. We extensively evaluate our method on three major benchmark datasets: KITTI (outdoor/autonomous driving), EuRoC-MAV (indoor/aerial drones), and TUM-RGBD (indoor/handheld), demonstrating state-of-the-art performance.
}



\maketitle

\renewcommand\thefootnote{}

\renewcommand\thefootnote{\fnsymbol{footnote}}
\setcounter{footnote}{1}

\section{Introduction}\label{sec1}

Monocular visual odometry (VO) plays a crucial role in autonomous systems by providing reliable pose estimates necessary for navigation \cite{li2017visual}, mapping \cite{huang2021survey,Deng_2024_CVPR,deng2024incremental,deng2025mne,10879467}, and scene understanding \cite{hua2023domain,deng2024multi} across a variety of platforms, from self‑driving vehicles and aerial drones to mobile robots and handheld devices.
Traditional geometry-based systems \cite{mur2017orb,xu2025airslam} deliver accurate localization in static, well-textured environments but suffer from error drift, failures in textureless or dynamic scenes, and sensitivity to lighting changes. Recent learning-based VO \cite{teed2021droid} has leveraged end-to-end networks to regress pose directly, improving robustness when supervised with high-precision ground-truth labels. To remove the need for costly annotations, self-supervised methods \cite{zhou2017unsupervised, bian2019unsupervised, zhao2023self} exploit photometric and geometric reconstruction losses between consecutive frames, enabling joint depth and ego-motion learning without explicit labels.

\begin{figure*}
    \centering
    \includegraphics[width=7in]{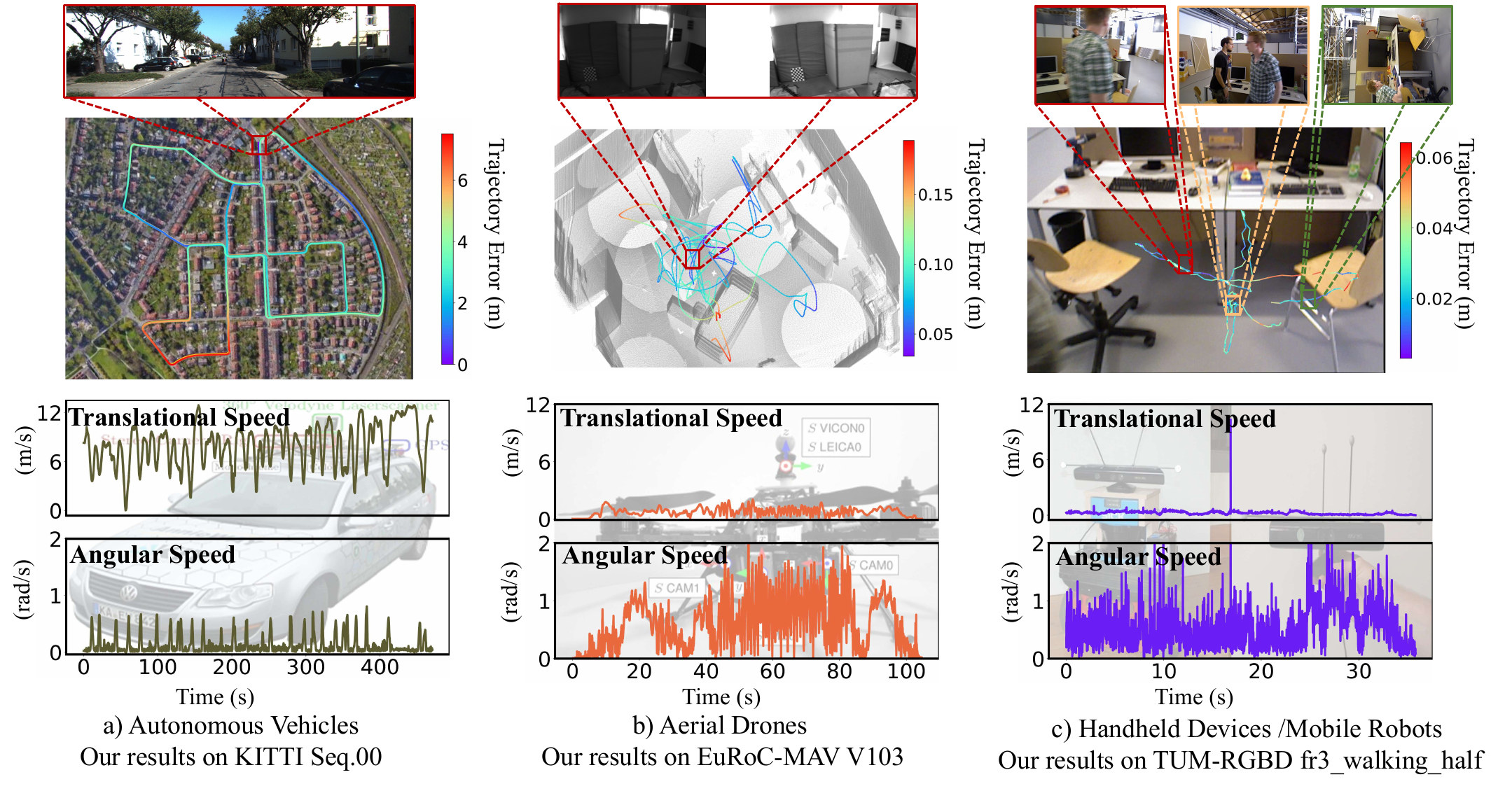}
    \caption{
    \textbf{Variations in environments and motion patterns across different deployment platforms.} 
    Trajectories estimated by our methods (\textbf{Middle}) in three representative datasets, with highlighting each platform’s characteristic scene features (\textbf{Top}): (a) autonomous vehicles (KITTI \cite{geiger2013vision}) features mostly planar, translation-dominant driving motions; (b) aerial drones (EuRoC \cite{burri2016euroc}) and (c) handheld devices (TUM-RGBD \cite{sturm12iros}) introduce complex 3D movements with rapid rotations, varying illumination, and dynamic occlusions.
    (\textbf{Bottom:}) Motion pattern distributions for each platform, plotting translational versus rotation speed to emphasize their distinct dynamics.
    }
    \label{Fig_1}
\end{figure*}

However, photometric and geometric losses provide sufficient constraints only in the smooth, translation-dominant motions of driving scenes and break down under more intricate movements \cite{bian2021auto}.
In Figure \ref{Fig_1}, for instance, aerial drones primarily challenge VO with rapid 6-DoF motion patterns and abrupt illumination changes, while handheld devices encounter severe dynamic occlusions. 
Under these conditions, a combination of rapid, rotation-dominated maneuvers and intermittent scene blockage, a simple pose model cannot reliably recover inter-frame poses, leading to significant drift and degraded accuracy.

To bridge this generalization gap, cutting-edge methods aim at improving the network’s ability to represent frame-to-frame ego-motion, what we refer to as \textbf{Capacity Enhancement}, rely on deeper architectures \cite{oquab2024dinov2,wang2024dust3r, zhao2023self}, regression paradigm (geometry-based \cite{zhan2021df, zhao2020towards} or learning-based \cite{zhao2023self}), or training strategy \cite{liu2024adaptive, bian2021auto} to capture diverse platform-specific characteristics. By expanding model expressiveness, these techniques can handle a wider range of motion patterns, yet the added parameters and training complexity often lead to brittle convergence and difficulty adapting online. In contrast, approaches centered on \textbf{Inter-Frame Tracking} seek to enforce global trajectory consistency by constructing pose graphs or applying multi-frame bundle adjustment (BA) \cite{yang2020d3vo,zou2020learning,teed2021droid}. These methods correct drift over long sequences but usually depend on hand-tuned keyframe selection and suffer when image pairs are far apart or contain severe occlusions. 
Conceptually, these two approaches correspond to the front-end (local state estimation) and back-end (global state optimization) of classical Simultaneous Localization and Mapping (SLAM) \cite{mur2017orb}.
However, capacity enhancement approaches address only local state estimation and inter-frame tracking approaches address only global state optimization, neither strategy alone endows a monocular VO system with robust performance across complex deployment such as aerial drones and handheld systems.

We therefore propose UNO, a unified framework that tightly couples adaptive local estimation, runtime model selection, and explicit geometric refinement within a single pipeline. 
First, our decoder-centric ego-motion estimation module, inspired by the Mixture-of-Experts \cite{zhou2022mixture}, deploys several lightweight decoders, each fine-tuned for a specific frame-to-frame motion patterns, to adaptively diverse deployment platforms.
Next, a fully differentiable Gumbel-Softmax selector module builds inter-frame correspondences at run time and activates the decoder whose learned expertise best aligns with the observed motion and scene conditions, replacing static heuristics with a data-driven gating mechanism. 
Finally, the selected decoder outputs and accompanying pseudo-depth priors are refined by a sliding-window bundle adjustment back-end that jointly optimizes camera poses and recovers global scale, all without external pose or depth supervision. By integrating adaptive local inference with this dynamic global refinement, UNO achieves state-of-the-art accuracy and reliable cross-domain generalization across a wide spectrum of deployment platforms.

Our main contributions can be summarized as follows:
\begin{itemize} 
    \item We propose a decoder-centric module for ego-motion estimation, in which multiple specialized decoders adapt to the distinct motion patterns of each deployment platform. This flexible multi‑decoder architecture overcomes the rigidity of single‑decoder designs and substantially enhances frame-to-frame local state estimation accuracy.
    \item We design a differentiable selection module based on the Gumbel-Softmax sampling trick to construct robust inter-frame associations. This method integrates explicit graph-based constraints with learnable tracking, enabling the joint optimization of inter-frame connectivity and decoder selection under varying motion conditions. 
    \item  We propose a lightweight bundle adjustment back‑end that operates on sparse inter-frame associations and uses a pre‑trained scale‑agnostic depth model to jointly refine global scale-consistent poses. By embedding affine depth recovery into this efficient optimization pipeline, our module delivers real‑time metric consistency and robust trajectory correction without any groundtruth supervision.
    \item We extensively validate our approach on three standard datasets, including KITTI \cite{geiger2013vision} (outdoor / autonomous vehicles), EuRoC-MAV \cite{burri2016euroc} (indoor / aerial drones), and TUM-RGBD \cite{schubert2018tum} (indoor / mobile robot and handheld devices), demonstrating its superior performance. 
\end{itemize}

\section{RELATED WORK}\label{sec2}

This paper addresses the challenges of self-supervised monocular VO across diverse environments, platforms, and motion patterns.
Prior research evolves along two axes: model capacity enhancement through network architectural or training strategies, and inter-frame tracking via frame correlation modeling to suppress error propagation.

\subsection{Local state estimation with capacity enhancement}
Methods under the capacity enhancement category leverage the network’s internal ability to model inter-frame relationships and thus accurately estimate the camera’s ego-motion $T_{t \to t+1}$.
Typically, these methods consist of two main components: a feature encoder $f_{enc}$ that extracts visual representations from input images, and a decoder $f_{dec}$ that estimates the ego-motion rotation $R$ and translation $t$ between consecutive frames $T = [ R, t ]$. The ego-motion poses are then accumulated over time to reconstruct the camera trajectory $\{P\}$, forming the foundation of the odometry process:
\begin{equation}
\left\{
\begin{aligned}
    & X_t = f_{enc}(I_t, I_{t+1}) \\
    & T_{t \to t+1} = f_{dec}(X_t) \\
    & \{P_i\}_{i=1}^{N} = P_1 \prod_{k=1}^{i-1}T_{k \to k+1}
\end{aligned}
\right.
\label{Eq_II_1}
\end{equation}
These methods can be broadly categorized into geometry-based and learning-based approaches.

\textbf{Geometry-based approaches.} In these methods, the feature encoder extracts robust visual representations from each frame, typically by detecting and matching keypoints \cite{lindenberger2023lightglue}, computing optical flow \cite{teed2020raft}, or by employing deep encoder models \cite{zhan2021df} to capture the inter-frame relationships. The pose decoder then utilizes these features to establish correspondences and, by exploiting epipolar geometry \cite{wang2025kpdepth}, estimates the relative motion between consecutive frames. However, a major limitation of these approaches is that they often lack an absolute scale. To mitigate this issue, some methods estimate temporal disparity maps and apply ICP frameworks to recover metric scale; unfortunately, such techniques frequently suffer in low-texture environments where feature extraction and disparity estimation are unreliable.

\textbf{Learning-based approaches.} 
In contrast, learning-based approaches directly infer inter-frame motion from data. These methods train a pose network to regress the relative motion between consecutive frames, eliminating the need for explicit feature matching and traditional geometric constraints. By leveraging large-scale datasets, these networks learn complex motion patterns and estimate poses in an end-to-end manner, though their performance depends on the labeled data quality. Recently, self-supervised methods \cite{zhou2017unsupervised, bian2019unsupervised, zhan2021df} have been proposed to jointly estimate depth and pose using inter-frame photometric loss, removing the reliance on ground-truth pose labels. However, these methods struggle in scenarios with limited motion constraints, leading to imprecise pose estimation. 
To mitigate these issues, various enhancements have been proposed, including rotation‑invariant preprocessing \cite{bian2021auto}, integration of foundation model depth priors \cite{sun2023sc}, attention‑based feature calibration \cite{wei2024fine, zhao2023self}, confidence‑aware flow estimation \cite{feng2024scipad}, and multi‑task shared‑parameter designs \cite{wei2024lite}.

\begin{figure*}
    \centering
    \includegraphics[width=7in]{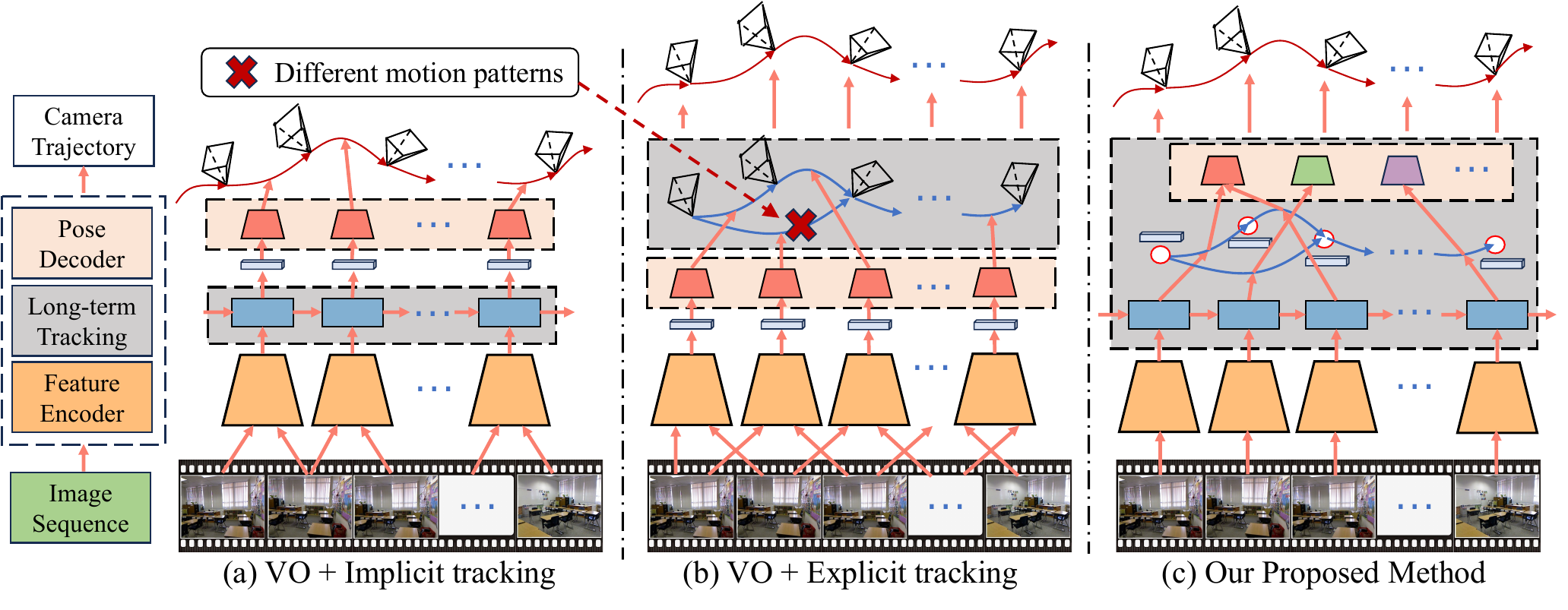}
    \caption{\textbf{Comparison of our method with existing approaches.} (a) Implicit motion modeling with long-term model, or (b) explicit tracking with pose graphs or bundle adjustment. In contrast, (c) our approach introduces a long-term model to implicitly capture feature correlations across frames, dynamically selecting different pose decoders for different motion patterns. This design enhances adaptability to diverse motion dynamics while ensuring robust pose estimation.}
    \label{Fig_2}
\end{figure*}

Despite extensive advances in both geometry-based and learning-based ego-motion estimation, current capacity-enhancement techniques remain constrained by their inability to prevent error accumulation arising from inter-frame constraints. Geometry-based pipelines depend on heavy keypoint matching and often lack absolute scale \cite{lindenberger2023lightglue}, while learning-based encoder incur large parameter overhead and struggle to generalize when photometric or motion priors are violated \cite{zhou2017unsupervised}. Even with enhancements such as rotation-invariant preprocessing \cite{bian2021auto} or foundation-model depth priors \cite{sun2023sc}, pose drift persists under rapid rotations, dynamic occlusions, and lighting changes. To address these challenges, we draw inspiration from Mixture-of-Experts \cite{zhou2022mixture} and foundational models like DINO \cite{oquab2024dinov2} and Dust3R \cite{wang2024dust3r} to propose a decoder-centric module that generates multiple candidate poses and selects the best estimate at run time. This design significantly reduces the number of parameters and enhances adaptability to diverse motion patterns across different deployment platforms.

\subsection{Global state optimization with inter-frame tracking}
As illustrated in Figure \ref{Fig_2}, cutting-edge methods with inter-frame constraint can be further divided into two groups: VO with implicit tracking, and VO with explicit tracking.

\textbf{Implicit tracking.} To mitigate the accumulated pose error drift inherent in VO, several works have introduced sequential models $f_{seq}$ to strengthen inter-frame constraints. In contrast to methods that solely rely on an encoder-decoder architecture to estimate poses from individual frame pairs (as in capacity enhancement approaches), these implicit tracking methods \cite{xue2019beyond, zou2020learning} integrate recurrent modules to capture the temporal evolution of features, thereby enabling more consistent long-term trajectory estimation:

\begin{equation}
\left\{
\begin{aligned}
    & X_t = f_{enc}(I_t, I_{t+1}) \\
    & Y_t, H_t = f_{seq}(X_t, H_{t-1}) \\
    & T_{t \to t+1} = f_{dec}(Y_t) \\
    & \{P_i\}_{i=1}^{N} = P_1 \prod_{k=1}^{i-1}T_{k \to k+1}
\end{aligned}
\right.
\label{Eq_II_2}
\end{equation}

Although these implicit methods integrate information from multiple frames to capture dynamic motion and scene changes, directly estimating non-Euclidean pose parameters with these networks remains extremely challenging. As a result, the pose estimates produced by implicit methods are often less accurate than those obtained through optimization-based explicit tracking approaches, which leverage direct geometric constraints to enforce consistency \cite{yang2020d3vo}.

\textbf{Explicit tracking.} 
Explicit tracking approaches enforce geometric consistency by constructing pose graphs or applying bundle adjustment after the initial pose estimation \cite{mur2017orb}.
In these approaches, keyframes $\Phi$ (selected by specific criteria or mechanisms) serve as nodes in the graph, and the relative ego-motion transformations between keyframes are encoded as edges. 
Finally, inter-frame optimization methods such as pose-graph optimization or sliding-window bundle adjustment refine the trajectory and correct the cumulative drift inherent in sequential pose estimation:

\begin{equation}
\left \{
\begin{aligned}
    & I_s, I_{t} = \Phi (\{I_i\}_{i=1}^N) \\
    & X_{s,t} = f_{enc}(I_s, I_{t})\\
    & T_{s \to t} = f_{dec}(X_t) \\
    & \{P_i^*\}_{i=1}^{N} = \arg\min_{\{P_i\}_{i=1}^{N}} \sum_{(i, j) \in \Phi} d(P_j, P_i \cdot T_{i \to j}^{-1})
\end{aligned}
\right.
\label{Eq_II_3}
\end{equation}

In recent years, some self-supervised methods have integrated explicit optimization processes such as bundle adjustment into the training pipeline \cite{lu2023deep, song2024graphavo, yang2020d3vo}. 
However, their accuracy remains fundamentally tied to the quality of initial frame-to-frame motion estimates, leading to failures under large inter-frame displacements, rapid rotations, lighting shifts, or dynamic occlusions. This shortfall stems from the decoupling of local state estimation and global state optimization: implicit tracking networks offer temporal smoothing without strict geometric enforcement, whereas explicit solvers apply bundle adjustment or pose-graph refinement only after independent feature matching and keyframe selection. Neither paradigm adapts its inference strategy to the current motion patterns, which degrades performance on platform-specific dynamics. To overcome these limitations, we propose an `encoder-tracking-decoder' architecture that embeds adaptive model selection directly into the optimization loop, fully leveraging inter-frame constraints for robust VO across diverse deployment scenarios.

\textbf{In summary}, existing self-supervised monocular VO methods either enhance local state estimation through specialized architectures (e.g., deep or recurrent networks) or enforce global state consistency via inter-frame tracking (e.g., pose graphs or bundle adjustment), but both rely on static decoders that cannot flexibly adapt to heterogeneous motion patterns \cite{zhao2020towards, zhao2023self}. Even recent pipelines that incorporate pseudo-depth priors (SC-DepthV3\cite{sun2023sc}, FrozenRecon\cite{xu2023frozenrecon}) leave pose regression and optimization decoupled, limiting robustness under rapid rotations or dynamic occlusions. To overcome these shortcomings, we generate multiple candidate poses per frame pair using a Mixture-of-Experts decoder ensemble and employ a differentiable Gumbel-Softmax selector to choose the most reliable estimate and impose strong inter-frame constraints. Moreover, our learnable sparse pose-graph in conjunction with multi-source inter-frame estimation facilitates on-the-fly data augmentation and constrains the optimization domain, enabling efficient global pose refinement without full-dataset access.

\section{Background and Overview}\label{sec3}

\subsection{Background and problem formulation}
Our framework takes a monocular image sequence $\{I_i\}_{i=1}^N$ and an initial camera pose $P_1$ as input, and it estimates the camera trajectory $\{P_i \in \textbf{SE}(3)\}_{i=1}^N$.
Similar to mainstream self-supervised approaches \cite{zhou2017unsupervised, sun2023sc, zhao2023self}, we utilize image depth to formulate photometric $L_{photo}$ and geometric $L_{geo}$ losses for pose training:
\begin{equation}
\label{Eq_III_1}
    L = \sum_{(i, j) \in \Phi}(\alpha L_{photo} + \beta L_{geo}),
\end{equation}
where $\alpha$ and $\beta$ are weights of each constraints, and $\Phi$ denotes the set of image pairs that satisfy the inter-frame constraint.

The photometric loss is defined as the inverse warping reconstruction error between the target frame $I_t$ and the reconstructed frame $\hat{I}_t$:
\begin{equation}
L_{\text{photo}} = 0.85 \frac{1 - \text{SSIM}(I_t, \hat{I}_t)}{2} + 0.15 \left\| I_t - \hat{I}_t \right\|_1,
\label{Eq_III_2}
\end{equation}
and the geometric loss is defined as the normalized difference between the point clouds projected from the depth maps \cite{zhao2023self}:
\begin{equation}
\label{eq13}
    L_{\text{geo}} = \frac{\| T_{i \to j} D_i(p) - D_j(p) \|}{T_{i \to j} D_i(p) + D_j(p)}.
\end{equation}\label{Eq_III_3}
where $p$ denotes a pixel coordinate in the image domain, and SSIM represents the structural similarity index. 

\subsection{Method overview}

\begin{figure*}
      \centering
	  \includegraphics[width=6in]{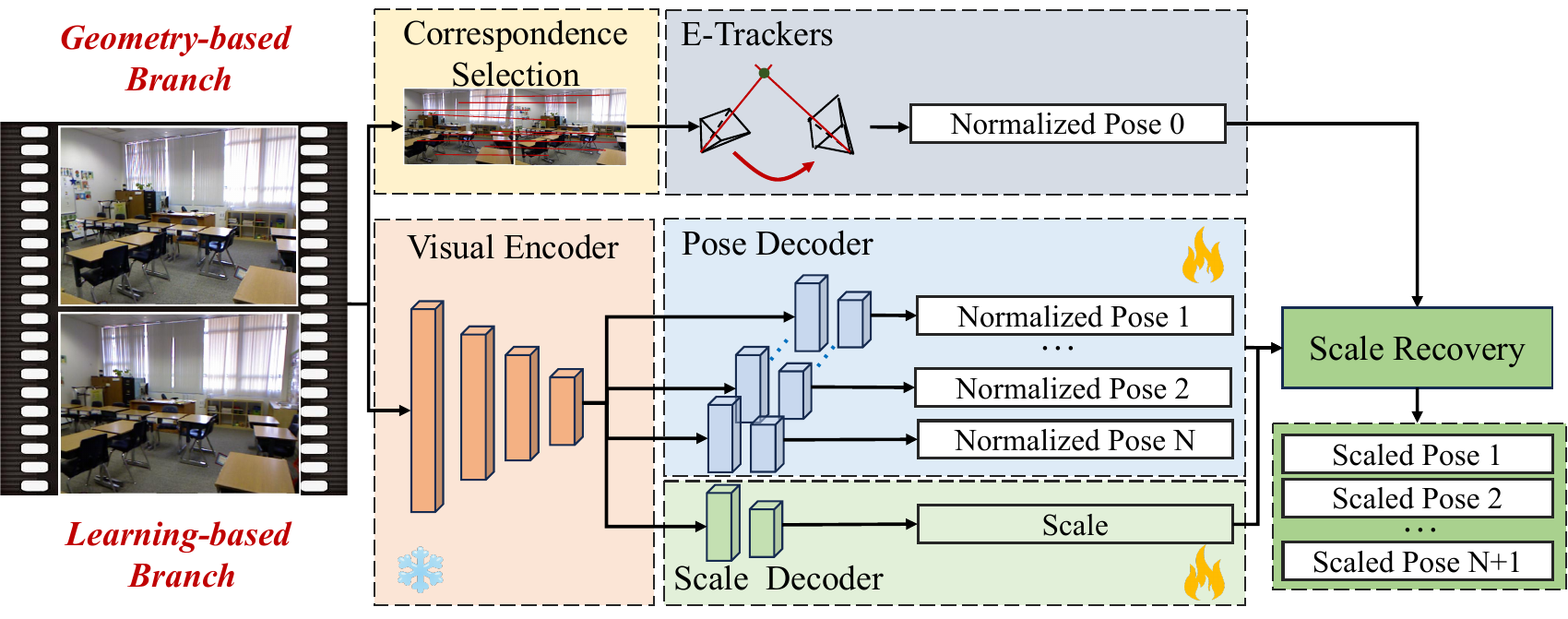}
      \caption{The framework of decoder-centric ego-motion estimation module.}
      \label{Fig_3}
\end{figure*}

In contrast to state-of-the-art self-supervised methods that are often tailored to specific motion patterns and struggle to generalize across diverse scenarios and deployment platforms, our framework is designed to handle arbitrary environments platforms, and motion variations, ensuring broader applicability and robustness. Our architecture is illustrated in Figure \ref{Fig_2}.

A key innovation of our framework is a decoder-centric local state estimation module that incorporates a mixture of experts (MoE) pose decoder (see \textbf{Section \ref{sec4}}). Drawing inspiration from prior work, we analyze the motion characteristics across different scenarios and the strengths and weaknesses of various pose estimation algorithms. Rather than investing heavily in complex feature extraction models, our approach focuses on training multiple lightweight decoders, which have significantly fewer parameters than the encoder. This design reduces computational overhead while enhancing the robustness of frame-to-frame local state estimation across a wide range of conditions.

Another critical component is our Gumbel sampling optimization, which integrates the advantages of long-term implicit multi-frame tracking and explicit tracking based on bundle adjustment and pose graphs (see \textbf{Section \ref{sec5}}). At the feature level, we construct explicit pose graph relationships and dynamically select the appropriate decoder to optimize the pose. Furthermore, instead of training a dedicated depth network, we utilize a large-scale pre-trained depth estimation model without fine-tuning. Although this model offers strong robustness across diverse environments, its output is affine-invariant and lacks absolute scale consistency. To overcome this limitation, we propose a sliding scale-agnostic bundle adjustment module that achieves globally consistent scale recovery and enhances global pose optimization.

\section{Local state estimation by decoder-centric module}\label{sec4}
We use the ego-motion estimate between two frames as the local state to be optimized.

\subsection{Rethinking the ego-motion estimation method}

Methods for estimating pose from two images (ego-motion) can be broadly classified into geometry-based and learning-based approaches, each offering distinct advantages and limitations.

\textbf{Epipolar Geometry:} These methods estimate pose from feature correspondences using the essential or fundamental matrix. While effective in well-structured environments, they suffer from \textit{motion degeneracy} (e.g., pure rotational motion renders translation recovery unsolvable) and \textit{structure degeneracy} (e.g., in planar scenes). Furthermore, epipolar geometry methods typically provide only scale-free translation estimates.

\textbf{Learning-Based Approaches:} These methods use deep networks to directly regress pose from images without relying on explicit geometric constraints. Although they bypass the need for feature matching and can be robust in many cases, their performance heavily depends on the training data distribution and they often struggle to generalize to unseen motion patterns.

Geometric methods focus on the quality of feature correspondences while often overlooking the differences in motion between frames. Although their rotation estimates are generally reliable, they tend to falter in challenging scenarios such as pure rotation or low-texture scenes. Conversely, learning-based methods tend to be more robust. However, the non-Euclidean nature of rotation makes gradient backpropagation challenging, resulting in mediocre rotational accuracy.

To overcome these limitations, we propose a decoder-centric ego-motion estimation module that dynamically selects specialized decoders to handle diverse motion patterns. Inspired by the success of Mixture-of-Experts (MoE) architectures \cite{zhou2022mixture}, our framework employs multiple expert decoders, such as geometry-based and learning-based methods, instead of a single decoder. A gating network dynamically activates the most relevant decoder based on real-time motion patterns. This design not only overcomes the inherent limitations of single-decoder architectures but also significantly improves ego-motion estimation accuracy by leveraging the complementary strengths of both geometric and learning-based approaches.

\begin{figure*}
      \centering
	  \includegraphics[width=6.5in]{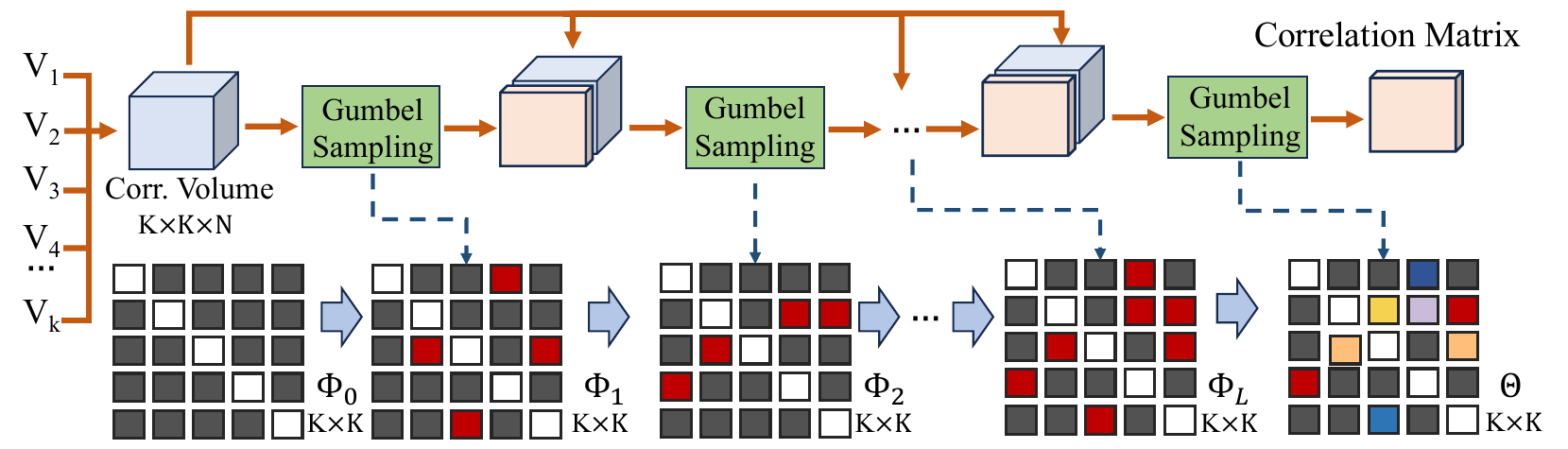}
      \caption{The architecture of sparse correlation graph construction and decoder adaptation module.}
      \label{Fig_4}
\end{figure*}

\subsection{Image encoder}

Traditional methods often stack two frames and apply convolution and pooling layers to detect differences. However, given the significant variation in robot motion across different scenes and platforms, such an approach may fail to capture the nuanced dynamics present in each frame. In our framework, each image is encoded individually using a frozen foundation encoder \cite{oquab2024dinov2} to extract stable feature representations:
\begin{equation}
\begin{aligned}
    & X_t = f_{enc}(I_t).
\end{aligned}
\label{Eq_IV_1}
\end{equation}

These per-frame features are then associated using a temporal RNN, which aggregates information from multiple frames to establish robust inter-frame relationships:
\begin{equation}
\label{Eq_IV_2}
\begin{aligned}
    & Y_t, H_t, \Phi_t, \Theta_t = RNN(X_t, H_{t-1}, \Phi_{t-1}).
\end{aligned}
\end{equation}
Here, $\Theta$ and $\Phi$ represent the modules for constructing explicit inter-frame constraints and for adapting the decoder, respectively. This design effectively accommodates the variability in motion across different environments and platforms, thereby enhancing the overall adaptability and robustness of the VO system. Further implementation details are provided in \textbf{Section \ref{sec5}}.

\subsection{Mixture-of-Experts (MoE) pose decoder}

To achieve robust ego-motion estimation, we design a Mixture-of-Experts (MoE) pose decoder module, as illustrated in Figure \ref{Fig_3}. The decoder architecture computes two distinct branches: one based on epipolar geometry and another via a learning-based decoder.
Since epipolar geometry can only estimate the normalized translation $t^{geo}_{norm}$, we recover scale information using the output from the learning-based branch.
The learning-based decoder is designed to decouple scale from pose by independently estimating a normalized translation $t^N_{norm}$ and a scale factor $s$, such that the final ego-motion candidate is given by
$[ R, s\ t_{norm} ]$.

The final local state $T$ is obtained by fusing these ego-motion outputs through a function $\Theta$ that dynamically selects the optimal decoder output for each frame pair based on confidence measures or predefined criteria. This process is formulated as:

\begin{equation}
\label{Eq_IV_3}
\begin{aligned}
    & T_{s \to t} = \sum_{k=1}^{K}1[\Theta(Y_s, Y_t)=k]f_{dec}^k(Y_s, Y_t),
\end{aligned}
\end{equation}
where $f_{dec}^k$ denotes the $k$-th specialized decoder in our multi-decoder framework.
By integrating these methods, our framework dynamically selects and fuses pose estimates, effectively balancing geometric constraints with learned priors to enhance accuracy and robustness across diverse motion scenarios.

\section{Gumbel Sampling-based Optimization}\label{sec5}
After obtaining the local state $T$, we introduce mechanisms to derive both the constraint module $\Phi$ and the decoder adaptation module $\Theta$.
These modules facilitate the dynamic selection of the most appropriate pose decoder and enable the construction of an effective pose optimization graph, which is then used to generate the refined trajectory $\{P^*\}$ while reducing the drift accumulated from ego-motion errors. The optimization process is defined as:
\begin{equation}
\label{Eq_V_1}
\begin{aligned}
    \{P_i^*\}_{i=1}^{N} = \arg\min_{\{P_i\}_{i=1}^{N}} \sum_{(i, j) \in \Phi} d(P_j, P_i \cdot T_{i \to j}^{-1}(T_i, Y_j, \Theta)).
\end{aligned}
\end{equation}

\subsection{Sparse correlation graph by Gumbel sampling}
We first explain how to construct a connected inter-frame correlation graph for optimization. Most current methods build a fully connected optimization graph, where a single mis-estimated pose can derail the entire framework. To this issue, we design an adaptive sampling method based on Gumbel-Softmax trick \cite{jang2017categorical} to generate a sparse inter-frame correlation graph. This sparse sampling reduces the optimization cost and avoids incorrect pairwise correspondences that could negatively impact pose estimation.

The initial pose-graph is denoted by $\Phi^0 \mathcal{(V,E)}$, where each vertex node $v_i \in \mathcal{V}$ represents each image feature and pose $P_i \in SE(3)$, and edge $(i,j) \in \mathcal{E}$ encodes the relative ego-motion pose $T_{ij}$ between image nodes.
We recursively sample $\Phi^0$ using a recurrent network to obtain the final $ \Phi^L = f_{samp} (\Phi^{L-1})$.

As shown in Figure \ref{Fig_4}, we first build a correlation volume $C \in \mathbb{R}^{K,K,F}$ by computing the dot product between all pairs of feature vectors within a sliding window of length $K$. And $F$ represents the feature dimension.
We then sample inter-frame relation on this correlation volume and last sampling result $\Phi^L = f_{samp}(C, \Phi^{L-1})$.
However, sampling following the Bernoulli distribution is discrete and non-differentiable, making it challenging for optimization through back-propagation.
To address this issue, we use a reparameterization trick, employing a Gumbel-Softmax scheme to generate pose graphs by sampling from the corresponding Gumbel-Softmax distribution \cite{mohamed2020monte}.

Specially, consider a categorical distribution where the probability of the $k_{th}$ class is denoted as $p_k$. Then, according to the Gumbel-Softmax sampling trick, we can generate a discrete sample $\hat{P}$ following the target distribution as follows:
\begin{equation}
\label{Eq_V_2}
    \hat{P} = \arg\max_k (log_{p_k} + g_k), k \in [0, 1],
\end{equation}
where $g_k=-log(-logU_k)$, and $U_k$ is sampled from a uniform distribution $U(0, 1)$.
The softmax function is applied to relax the argmax operation to obtain a real-valued vector $\tilde{P}$ by a differentiable function as in:
\begin{equation}
    \tilde{P}_k = \frac{exp((logp_k+g_k) / \tau)}{\sum_{j=0}^{k-1}exp((logp_j+g_j) / \tau)},
\end{equation}
where $\tau$ is a temperature parameter and $k$ is categories.

Finally, we further apply another Gumbel-Softmax sampling layer on the final layer's output and the correlation volume to obtain $\Theta$, which is used to select the most suitable decoder. Unlike the binary decision process used for constructing inter-frame associations, this selection process involves a categorical distribution over $k$ decoder clasees.
This differentiable sampling mechanism enables us to obtain a differentiable approximation of the discrete decoder choice, facilitating joint optimization.

\subsection{Pose refinement with scale-agnostic depth}

Existing self-supervised methods typically require joint refinement of both pose and depth as the quality of the photometric loss by inverse warping:
\begin{equation}
    \hat{I}_t = \pi(K T_{t \to s} \textbf{D}_t(x) K^{-1} \pi^{-1}(x)),
    \label{Eq_V_4}
\end{equation}
where $K$ is the camera intrinsic matrix and $\textbf{D}_t(x)$ represents the estimated depth at pixel \( x \) in frame \( I_t \).
The efficacy of pose optimization strongly depends on accurate depth estimation.
When depth is poorly estimated, the reconstruction error becomes unreliable, which hampers effective pose optimization. Moreover, depth networks trained on a single dataset often fail to generalize to diverse scenes.
To address these issues, we leverage recently proposed depth estimation foundation models that exhibit robust cross-domain performance. However, these models provide only affine-invariant depth, lacking the metric scale necessary for direct pose optimization. Our approach overcomes this limitation by integrating a sliding window bundle adjustment module that jointly optimizes the affine parameters (scale and shift) of the pseudo-depth. This enables us to effectively use the affine depth in our photometric optimization framework, thereby achieving more accurate pose estimation even in challenging, diverse environments.

\begin{algorithm}
\caption{Algorithm of our method}
\begin{algorithmic}[1]
\Require Sequence of images $\{I_i\}_{i=1}^N$ and initial pose $P_1$
\Ensure Optimized keyframe poses $\{P_i^*\}_{i=1}^{N}$
\State Encoding: $Y_i = f_{enc}(I_i)$
\State Correlation sampling: $\Phi = f_{samp}(X_i)$
\State Decoder sampling: $\Theta = g_{samp}(X_i, \Phi)$
\ForAll {$(I_i, I_j)$ in $\Phi$}
    \State Local estimation:                
    $T_{i \to j} = (1[\Theta_{i,j}=k]f_{dec})^k(Y_i, Y_j)$
\EndFor
\Statex \# \textbf{Backend refinement.}
    \For {iter to Iterations}
        \State $\hat{D} = f_{depth}(I_i)$
        \State Sample point pair $(p_i, p_j) \leftarrow f_{geo}$.
        \State $A_i, B_i = f_{LWLR}(D_i, p_i, p_j)$
        \State $D_i = A_i \odot \hat{D}_i + B_i$
        \State Backward warp $I_i$ to $\Phi$
        \State Back-propagate $T_{i\to j}$, $A$ and $B$ with Eqn. \ref{Eq_III_1}.
    \EndFor
\State $P^* = P_1 \prod_{k=1}^{i-1}P_k^{k+1}$
\State Back-propagate $f_{dec}, f_{samp}, g_{samp}$ with Eqn. \ref{Eq_V_1}.
\end{algorithmic}
\label{algor_1}
\end{algorithm}

We model the relationship between affine-invariant depth $\hat{D}$ and metric depth $\textbf{D}$ using a linear transformation:

\begin{equation}
\label{Eq_IV_9}
    \textbf{D} = \textbf{A} \odot \hat{\textbf{D}}+\textbf{B},
\end{equation}
where $\textbf{A}$ is the scale factor and $\textbf{B}$ is the translation offset. These parameters capture the affine transformation needed to convert the affine-invariant output of our pre-trained depth model into a metric depth. Following a strategy similar to FrozenRecon \cite{xu2023frozenrecon}, we jointly optimize $A$ and $B$ within our sliding window bundle adjustment module, ensuring that the adjusted depth values contribute effectively to a reliable photometric loss and ultimately improve pose estimation accuracy.

Unlike FrozenRecon, our approach does not rely on uniform image sampling to construct the sliding window bundle adjustment. Instead, we leverage matching-based rigid feature associations derived from epipolar geometry with RANSAC filtering to obtain reliable rigid transformations between corresponding frames:
\begin{equation}
\label{Eq_IV_10}
    \textbf{A}, \textbf{B} = f_{LWLR}(\sum_{x,y \in \Omega}(\textbf{d}_j(x), \{ \omega_{i,j}\textbf{d}_i(y) \})).
\end{equation}
These filtered transformations form the basis of our optimization graph, ensuring that only geometrically consistent regions contribute to the depth-to-pose conversion. In outdoor scenes such as those in KITTI, uniform sampling is often disrupted by non-informative regions like the sky and by dynamic objects that degrade accuracy. By focusing on robust, matching-based correspondences, our method effectively mitigates these issues and significantly enhances overall pose estimation accuracy.
The optimization pipeline is summarized in Algorithm \ref{algor_1}.

\section{Experiments}\label{sec6}

Our experimental evaluation is designed to rigorously validate the performance and generalization ability of our self-supervised pose estimation framework across a wide range of platforms and scenes. First, we detail the implementation specifics and overall experimental setup. Secondly, we benchmark our method on three standard datasets (Figure \ref{Fig_1}): 
\begin{itemize} 
\item \textbf{KITTI} \cite{geiger2013vision}, which primarily captures smooth, planar vehicle motion at high speeds, making it ideal for testing long-range translational accuracy in outdoor driving scenarios. 
\item \textbf{EuRoC-MAV} \cite{burri2016euroc}, which presents challenging 6‑DoF trajectories with rapid and intense motion patterns typical of aerial drone, thus providing an extreme case for evaluating pose estimation in dynamic aerial environments. 
\item \textbf{TUM RGB‑D} \cite{sturm12iros}, which records complex, dynamic rotations captured by handheld devices and mobile robotic platforms in indoor environments. These sequences are characterized by frequent occlusions, abrupt changes in viewpoint, and crowded scenes, challenging the framework's ability to maintain accurate pose estimation under cluttered conditions.
\end{itemize}
Covering scenarios from smooth translational driving to rapid, extreme aerial and complex indoor rotations, these datasets constitute a comprehensive testbed to demonstrate the robustness and adaptability of our approach.
Finally, we present extensive ablation and interpretability experiments to further substantiate the effectiveness of our method.

\subsection{Implementation details}
Our method is implemented based on the Pytorch framework. First, we extract features from each frame using the DINOv2 foundation model \cite{oquab2024dinov2}. Our Mixture-of-Experts decoders consist of a Epipolar geometry decoder and seven lightweight learning-based decoders.
Each learning-based decoder follows the Sc-Sfmlearner architecture \cite{bian2019unsupervised}, which are dynamically selected based on motion characteristics. The geometry branch detects and matches SIFT features and solves the essential matrix via a GPU‑accelerated five‑point algorithm \cite{zhu2023lighteddepth}, producing normalized translations. The pose refining module uses LeRes \cite{yin2021learning} to obtain scale‑agnostic depth estimates, which are then refined via our bundle adjustment module.

For training, we adopt the Adam optimizer with a learning rate of $10^{-4}$, $\beta_1 = 0.9$ and $\beta_2 = 0.999$. The weights for loss functions in Eqn. \ref{Eq_III_1} are empirically set to $\alpha = 1$ and $\beta = 0.1$ We train the network for $200$ epochs with $1000$ sampled batches per epoch and validate the model at the end of each epoch.
Horizontal flipping and color jitter augmentation is incorporated during training with a probability of $0.1$.
For validation, we adopt the real‐scale alignment technique of Sc-Sfmlearner \cite{bian2019unsupervised}, ensuring that all methods are evaluated under a consistent metric scale. 

\begin{figure*}
      \centering
	  \includegraphics[width=7in]{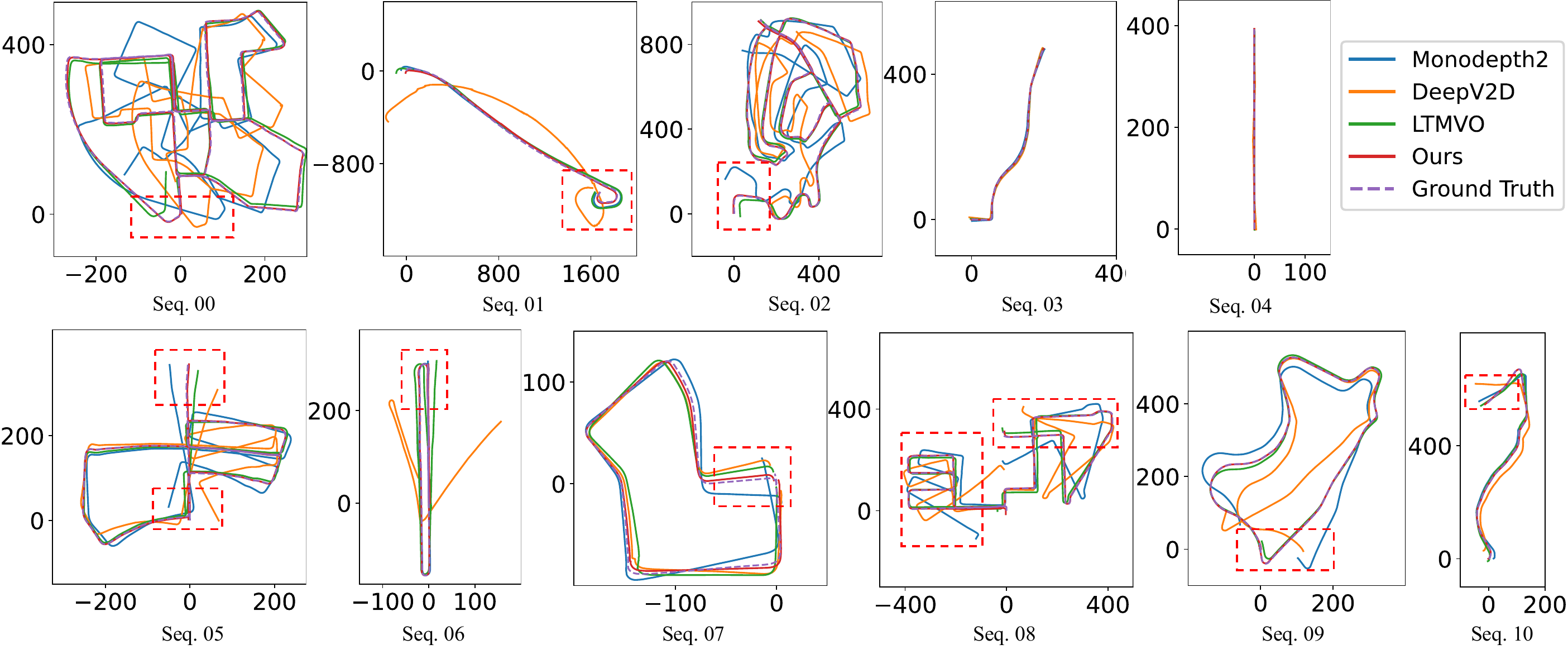}
      \caption{Qualitative results on the KITTI odometry benchmark.
      }
      \label{Fig_5}
\end{figure*}

\begin{table}
\centering
\caption{Odometry results on the KITTI Odometry test split. 
Best results are highlighted in \colorbox{Min1}{\textbf{first}}, \colorbox{Min2}{\underline{second}}, and \colorbox{Min3}{\textit{third}}, respectively.}
\begin{tabular}{c|cccc}
\hline
                         & \multicolumn{2}{c}{Seq. 09}                                 & \multicolumn{2}{c}{Seq. 10}                                 \\
\multirow{-2}{*}{Method} & $T_{err} \downarrow$                       & $R_{err} \downarrow$                       & $T_{err} \downarrow$                       & $R_{err} \downarrow$                       \\ \hline
Sfmlearner \cite{zhou2017unsupervised}               & 8.28                         & 4.07                         & 15.25                        & 4.06                         \\
Sc-Sfmlearner \cite{bian2019unsupervised}            & 7.38                         & 3.04                         & 7.65                         & 4.85                         \\
LTMVO \cite{zou2020learning}                   & 3.49                         & 1.03                         & 5.81                         & 1.82                         \\
DF‐VO‐M \cite{zhan2021df}                 & 2.40                         & \cellcolor{Min2}\underline{0.24} & \cellcolor{Min3}\textit{1.82} & \cellcolor{Min3}\textit{0.38} \\
MLF-VO \cite{jiang2022self}                  & 3.90                         & 1.41                         & 4.88                         & 1.38                         \\
GraphAVO \cite{song2024graphavo}                & 5.87                         & 1.53                         & 9.07                         & 3.63                         \\
Lite-SVO \cite{wei2024lite}                & 7.20                         & 1.50                         & 8.50                         & 2.40                         \\
SCIPaD \cite{feng2024scipad}                  & 7.43                         & 2.46                         & 9.82                         & 3.87                         \\
Fine-MVO \cite{wei2024fine}                & 3.37                         & 1.00                         & 5.34                         & 1.20                         \\
AdaptiveHVO \cite{liu2024adaptive}             & \cellcolor{Min3}\textit{2.01} & 1.35                         & 2.01                         & 1.32                         \\
KPDepth-VO \cite{wang2025kpdepth}              & \cellcolor{Min2}\underline{1.52} & \cellcolor{Min3}\textit{0.37} & \cellcolor{Min2}\underline{1.70} & \cellcolor{Min2}\underline{0.45} \\ \hline
Ours                     & \cellcolor{Min1}\textbf{0.69}& \cellcolor{Min1}\textbf{0.22} & \cellcolor{Min1}\textbf{0.61} & \cellcolor{Min1}\textbf{0.25} \\
& \color{red}{$54.6 \% \uparrow $} & \color{red}{$8.3 \% \uparrow $} & \color{red}{$64.1 \% \uparrow $} & \color{red}{$44.4 \% \uparrow $} \\ 
\hline
\end{tabular}
\label{Tab_2-1}
\end{table}

\begin{table*}[htbp]
\centering
\caption{Odometry results on the KITTI Odometry Dataset Seq. 00-10.
$\dagger$ trained on Seq. 00, 01, 02, 08, and 09 split, and $\ddagger$ use Eigen split.
Best results are highlighted in \colorbox{Min1}{\textbf{first}}, \colorbox{Min2}{\underline{second}}, and \colorbox{Min3}{\textit{third}}, respectively.}
\begin{tabular}{cc|c|ccccccccccc}
\hline
\multicolumn{2}{c|}{Method}                                                                                                                                              & Metric    & 00                                    & 01                                    & 02                                    & 03                                    & 04                                    & 05                                    & 06                                    & 07                                    & 08                                    & 09                                    & 10                                    \\ \hline
\multicolumn{1}{c|}{\multirow{14}{*}{\rotatebox{90}{Capacity Enhancement}}} & \multirow{2}{*}{Sfmlearner \cite{zhou2017unsupervised}}     & $T_{err}$ & 19.28                                 & 21.71                                 & 18.99                                 & 9.73                                  & 3.17                                  & 10.03                                 & 11.00                                 & 11.69                                 & 8.67                                  & 8.28                                  & 12.20                                 \\
\multicolumn{1}{c|}{}                                                                          &                                                                         & $R_{err}$ & 5.67                                  & 2.59                                  & 3.33                                  & 3.50                                  & 3.26                                  & 3.60                                  & 3.78                                  & 5.90                                  & 2.56                                  & 3.07                                  & 2.96                                  \\
\multicolumn{1}{c|}{}                                                                          & \multirow{2}{*}{Sc-Sfmlearner \cite{bian2019unsupervised}} & $T_{err}$ & 6.92                                  & 86.41                                 & 6.15                                  & 5.12                                  & 4.14                                  & 6.07                                  & 5.91                                  & 7.64                                  & 6.79                                  & 7.38                                  & 7.65                                  \\
\multicolumn{1}{c|}{}                                                                          &                                                                         & $R_{err}$ & 2.92                                  & 1.22                                  & 2.70                                  & 4.22                                  & 2.77                                  & 2.99                                  & 2.81                                  & 4.04                                  & 3.06                                  & 3.04                                  & 4.85                                  \\
\multicolumn{1}{c|}{}                                                                          & \multirow{2}{*}{DF-VO-M \cite{zhan2021df}}                                                & $T_{err}$ & \cellcolor{Min3}\textit{2.33} & 39.46                                 & 3.24                                  & 2.21                                  & 1.43                                  & \cellcolor{Min3}\textit{1.09} & 1.15                                  & \cellcolor{Min1}\textbf{0.63} & 2.18                                  & 2.40                                  & 1.82                                  \\
\multicolumn{1}{c|}{}                                                                          &                                                                         & $R_{err}$ & 0.64                                  & 0.50                                  & 0.49                                  & 0.38                                  & 0.30                                  & \cellcolor{Min2}\underline{0.25} & 0.39                                  & \cellcolor{Min2}\underline{0.29} & 0.32                                  & \cellcolor{Min2}\underline{0.24} & 0.38                                  \\
\multicolumn{1}{c|}{}                                                                          & \multirow{2}{*}{SCIPaD \cite{feng2024scipad}}                                                 & $T_{err}$ & 6.19                                  & 10.39                                 & 4.82                                  & 4.84                                  & 2.74                                  & 3.78                                  & 3.93                                  & 5.64                                  & 4.67                                  & 7.43                                  & 9.82                                  \\
\multicolumn{1}{c|}{}                                                                          &                                                                         & $R_{err}$ & 2.26                                  & 1.17                                  & 1.67                                  & 1.63                                  & 3.84                                  & 1.75                                  & 1.32                                  & 4.04                                  & 1.85                                  & 2.46                                  & 3.87                                  \\
\multicolumn{1}{c|}{}                                                                          & \multirow{2}{*}{MLF-VO \cite{jiang2022self}}                                                 & $T_{err}$ & 2.61                                  & \cellcolor{Min3}\textit{7.57} & 4.32                                  & 2.68                                  & 2.70                                  & 2.35                                  & 1.61                                  & 5.49                                  & 3.15                                  & 3.92                                  & 4.90                                  \\
\multicolumn{1}{c|}{}                                                                          &                                                                         & $R_{err}$ & 0.51                                  & 0.55                                  & 0.66                                  & 0.75                                  & 0.75                                  & 0.60                                  & 0.46                                  & 2.68                                  & 0.64                                  & 1.41                                  & 1.36                                  \\
\multicolumn{1}{c|}{}                                                                          & \multirow{2}{*}{AdaptiveHVO \cite{liu2024adaptive}}                                            & $T_{err}$ & 3.32                                  & 9.96                                  & 3.57                                  & 3.96                                  & 2.64                                  & 4.20                                  & 3.64                                  & 2.53                                  & 2.99                                  & 2.01                                  & 2.01                                  \\
\multicolumn{1}{c|}{}                                                                          &                                                                         & $R_{err}$ & 1.39                                  & 1.78                                  & 1.93                                  & 3.32                                  & 3.54                                  & 1.72                                  & 1.88                                  & 1.46                                  & 1.40                                  & 1.35                                  & 1.32                                  \\
\multicolumn{1}{c|}{}                                                                          & \multirow{2}{*}{KPDepth-VO \cite{wang2025kpdepth}}                                             & $T_{err}$ & \cellcolor{Min2}\underline{1.76} & 12.64                                 & 3.14                                  & 2.49                                  & \cellcolor{Min2}\underline{0.70} & \cellcolor{Min2}\underline{0.96} & \cellcolor{Min3}\textit{0.77} & 2.57                                  & \cellcolor{Min3}\textit{1.49} & \cellcolor{Min3}\textit{1.64} & \cellcolor{Min3}\textit{1.52} \\
\multicolumn{1}{c|}{}                                                                          &                                                                         & $R_{err}$ & 0.60                                  & 0.48                                  & 0.59                                  & 0.41                                  & \cellcolor{Min2}\underline{0.26} & 0.34                                  & 0.27                                  & 0.99                                  & 0.38                                  & 0.28                                  & 0.37                                  \\ \hline
\multicolumn{1}{c|}{\multirow{10}{*}{\rotatebox{90}{Inter-frame Tracking}}} & \multirow{2}{*}{ORB-SLAM2 (w/o LC) \cite{mur2017orb}}                                     & $T_{err}$ & 11.43                                 & 107.57                                & 10.34                                 & \cellcolor{Min3}\textit{0.97} & \cellcolor{Min3}\textit{1.30} & 9.04                                  & 14.56                                 & 9.77                                  & 11.46                                 & 9.30                                  & 2.57                                  \\
\multicolumn{1}{c|}{}                                                                          &                                                                         & $R_{err}$ & 0.58                                  & 0.89                                  & \cellcolor{Min1}\textbf{0.26} & \cellcolor{Min1}\textbf{0.19} & \cellcolor{Min3}\textit{0.27} & \cellcolor{Min3}\textit{0.26} & 0.26                                  & \cellcolor{Min3}\textit{0.36} & \cellcolor{Min1}\textbf{0.28} & 0.26                                  & \cellcolor{Min3}\textit{0.32} \\
\multicolumn{1}{c|}{}                                                                          & \multirow{2}{*}{ORB-SLAM2 (w/ LC) \cite{mur2017orb}}                                      & $T_{err}$ & 2.35                                  & 109.10                                & 3.32                                  & \cellcolor{Min2}\underline{0.91} & 1.56                                  & 1.84                                  & 4.99                                  & \cellcolor{Min3}\textit{1.91} & 9.41                                  & 2.88                                  & 3.30                                  \\
\multicolumn{1}{c|}{}                                                                          &                                                                         & $R_{err}$ & \cellcolor{Min2}\underline{0.35} & \cellcolor{Min3}\textit{0.45} & \cellcolor{Min3}\textit{0.31} & \cellcolor{Min1}\textbf{0.19} & \cellcolor{Min3}\textit{0.27} & \cellcolor{Min1}\textbf{0.20} & \cellcolor{Min1}\textbf{0.23} & \cellcolor{Min1}\textbf{0.28} & \cellcolor{Min3}\textit{0.30} & 0.25                                  & \cellcolor{Min2}\underline{0.30} \\
\multicolumn{1}{c|}{}                                                                          & \multirow{2}{*}{BeyondTracking $\dagger$ \cite{xue2019beyond}}                               & $T_{err}$ & ‐                                     & ‐                                     & ‐                                     & 3.32                                  & 2.96                                  & 2.59                                  & 4.93                                  & 3.07                                  & ‐                                     & ‐                                     & 3.94                                  \\
\multicolumn{1}{c|}{}                                                                          &                                                                         & $R_{err}$ & ‐                                     & ‐                                     & ‐                                     & 2.10                                  & 1.76                                  & 1.25                                  & 1.90                                  & 1.76                                  & ‐                                     & -                                     & 1.72                                  \\
\multicolumn{1}{c|}{}                                                                          & \multirow{2}{*}{LTMVO \cite{zou2020learning}}                                                  & $T_{err}$ & 2.60                                  & 13.27                                 & \cellcolor{Min3}\textit{2.49} & 1.59                                  & 2.52                                  & 2.63                                  & 2.64                                  & 6.43                                  & 3.61                                  & 3.49                                  & 5.81                                  \\
\multicolumn{1}{c|}{}                                                                          &                                                                         & $R_{err}$ & \cellcolor{Min3}\textit{0.55} & \cellcolor{Min2}\underline{0.35} & 0.33                                  & 0.65                                  & 0.46                                  & 0.50                                  & 0.66                                  & 2.09                                  & 0.35                                  & 1.03                                  & 1.82                                  \\
\multicolumn{1}{c|}{}                                                                          & \multirow{2}{*}{D3VO $\ddagger$ \cite{yang2020d3vo}}                                        & $T_{err}$ & ‐                                     & \cellcolor{Min2}\underline{1.07} & \cellcolor{Min1}\textbf{0.80} & ‐                                     & ‐                                     & ‐                                     & \cellcolor{Min2}\underline{0.67} & ‐                                     & \cellcolor{Min1}\textbf{1.00} & \cellcolor{Min2}\underline{0.78} & \cellcolor{Min2}\underline{0.62} \\
\multicolumn{1}{c|}{}                                                                          &                                                                         & $R_{err}$ & ‐                                     & ‐                                     & ‐                                     & ‐                                     & ‐                                     & ‐                                     & ‐                                     & ‐                                     & ‐                                     & ‐                                     & ‐                                     \\ \hline
\multicolumn{1}{c|}{}                                                                          & \multirow{2}{*}{Ours}                                                   & $T_{err}$ & \cellcolor{Min1}\textbf{0.72} & \cellcolor{Min1}\textbf{1.05} & \cellcolor{Min2}\underline{0.89} & \cellcolor{Min1}\textbf{0.76} & \cellcolor{Min1}\textbf{0.34} & \cellcolor{Min1}\textbf{0.51} & \cellcolor{Min1}\textbf{0.61} & \cellcolor{Min2}\underline{1.54} & \cellcolor{Min2}\underline{1.06} & \cellcolor{Min1}\textbf{0.69} & \cellcolor{Min1}\textbf{0.61} \\
\multicolumn{1}{c|}{}                                                                          &                                                                         & $R_{err}$ & \cellcolor{Min1}\textbf{0.28} & \cellcolor{Min1}\textbf{0.28} & \cellcolor{Min1}\textbf{0.26} & \cellcolor{Min1}\textbf{0.19} & \cellcolor{Min1}\textbf{0.13} & 0.33                                  & \cellcolor{Min2}\underline{0.24} & 0.80                                  & \cellcolor{Min3}\textit{0.31} & \cellcolor{Min1}\textbf{0.22} & \cellcolor{Min1}\textbf{0.25} \\ \hline
\end{tabular}
\label{Tab_2}
\end{table*}

\subsection{Visual odometry results on public datasets}
\textbf{KITTI: outdoor autonomous driving scenarios.}
KITTI is widely recognized as an authoritative benchmark for autonomous driving scenarios. Collected from vehicles in urban and highway environments, it provides high-quality ground-truth data essential for evaluating VO and SLAM systems. Since the motion captured in KITTI is predominantly planar, with vehicles mainly experiencing high-speed translational movement along flat roads and minimal vertical or rotational dynamics, it serves as an ideal testbed for methods designed to handle autonomous driving scenes.

The KITTI Odometry Benchmark is a dedicated subset of the KITTI suite comprising eleven stereo sequences (Sequences 00–10) with publicly available 6‑DoF ground‑truth trajectories for VO evaluation. Following standard practice \cite{zhou2017unsupervised, bian2019unsupervised, wang2025kpdepth}, we train on Sequences 00–08 and evaluate on Sequences 09–10.
Table \ref{Tab_2-1} compares average translational error ($T_{err}, \%$) and rotational error ($R_{err}, \textdegree /100m$) for leading self-supervised methods on test split.
Our method reduces translational error to $0.69\%$ percent on Sequence 09 and $0.61\%$ percent on Sequence 10, and lowers rotation error to $0.22\textdegree/100m$ and $0.25\textdegree/100m$ respectively. This corresponds to more than a $50 \%$ improvement over the previous best method (KPD-Depth \cite{wang2025kpdepth}).

We then extend evaluation to all eleven sequences and categorize baselines into two groups: capacity enhancement and inter‑frame tracking on Table \ref{Tab_2}.
Tracking‑based approaches generally surpass recent capacity enhancement methods across most sequences.
It should be noted that BeyondTracking \cite{xue2019beyond} and D3VO \cite{yang2020d3vo} employ different dataset partitioning strategies - particularly, D3VO \cite{yang2020d3vo} utilizes the full KITTI Eigen split training set which contains substantially more training data (28,108 frames) compared to our odometry split (4,487 frames). More critically, their training set overlaps with sequences in our test split, potentially introducing test data leakage.
Nevertheless, our method achieves the top result on nearly every sequence, with the largest advantage on Sequence 01—the only sequence where vehicle speed exceeds $60$ km/h—highlighting the importance of motion‑aware decoder specialization and strong inter‑frame constraints for the large variation of motion patterns.
Figure \ref{Fig_5} shows the corresponding camera trajectories, further demonstrating the superior performance of our method.

\begin{figure*}
      \centering
	  \includegraphics[width=5.5in]{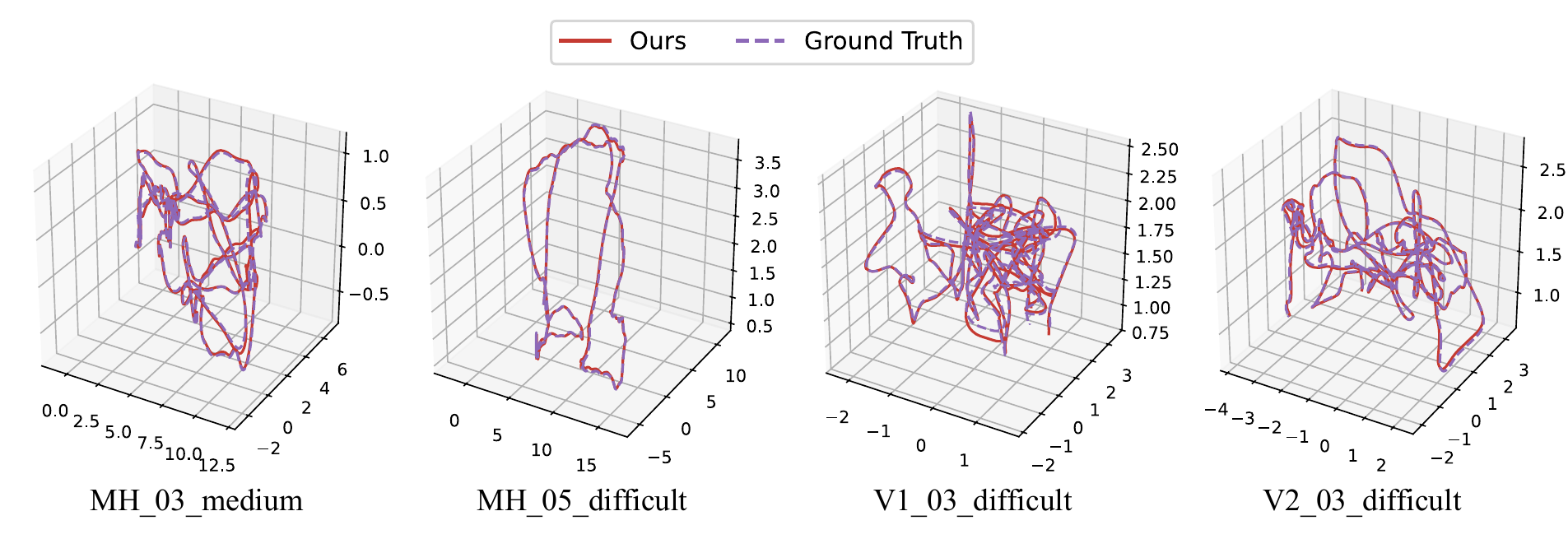}
      \caption{Qualitative results on test split \cite{yang2020d3vo} from the EuRoC.
      }
      \label{Fig_6}
\end{figure*}

\textbf{EuRoC-MAV: Indoor aerial drones.}
The EuRoC-MAV dataset presents an exceptionally challenging platform for VO, featuring aggressive camera motions, rapid rotations, and illumination inherent to aerial drones. Unlike ground vehicles' predominantly planar, high‑speed translational motion, EuRoC-MAV exposes algorithms to aggressive maneuvers and rapid viewpoint changes, which can cause motion blur and break photometric consistency.
Most existing self‑supervised methods struggle to maintain robustness under these conditions and often fail to track consistently over long trajectories.

Recognizing that many self‑supervised VO approaches cannot accurately estimate aerial drones motion, in addition to learning-based methods (CC \cite{ranjan2019competitive} and D3VO\cite{yang2020d3vo}), we also supplement classical geometry‑based VO (ORB-SLAM2\cite{mur2017orb} and DSO \cite{engel2017direct} and visual‑inertial odometry (VIO) approaches (VINS \cite{qin2018vins}, OKVIS \cite{leutenegger2015keyframe}, ROVIO \cite{bloesch2015robust}, MSCKF \cite{mourikis2007multi}, and SVO \cite{forster2016manifold}). 
Following the D3VO split \cite{yang2020d3vo}, we reserve the sequences MH\_03\_medium, MH\_05\_difficult, V1\_03\_difficult, V2\_02\_medium, and V2\_03\_difficult as our test set, using the remaining sequences for training.

Table \ref{Tab_3} shows the Absolute Trajectory Error (ATE), defined as the root‑mean‑square of the translational deviations after trajectory alignment, on all test sequences excluded V2\_03\_difficult (V2\_03\_difficult is usually excluded due to many missing images from one of the cameras).
Our method consistently outperforms existing approaches across most sequences, underscoring its robustness under aggressive dynamic motion and extreme viewpoint changes with rapid rotations.
Moreover, on several sequences, our self‑supervised method even surpasses visual‑inertial baselines, demonstrating competitive performance without relying on IMU data.
Figure \ref{Fig_6} illustrates the estimated trajectories, further demonstrating the superior accuracy of our approach.

\textbf{TUM-RGBD: Indoor handheld devices and mobile robots.}
The TUM RGB‑D dataset is a well‑established benchmark for indoor VO evaluation, captured with mobile robot and handheld devices in dynamic environments.
Its sequences encompass cluttered indoor scenes where dynamic objects frequently interfere with camera motion, breaking photometric assumptions and challenging VO robustness.

Following established evaluation protocols from BeyondTracking \cite{xue2019beyond} and LTMVO \cite{zou2020learning}, we implement the same dataset split to ensure methodological consistency. 
To comprehensively assess performance variations across different training strategies, we conduct two configurations of SC-DepthV3 \cite{sun2023sc}: (1) using their official weights trained with alternative splits, and (2) our re-implementation strictly following our split protocol.

\begin{figure}
      \centering
        \includegraphics[width=3.35in]{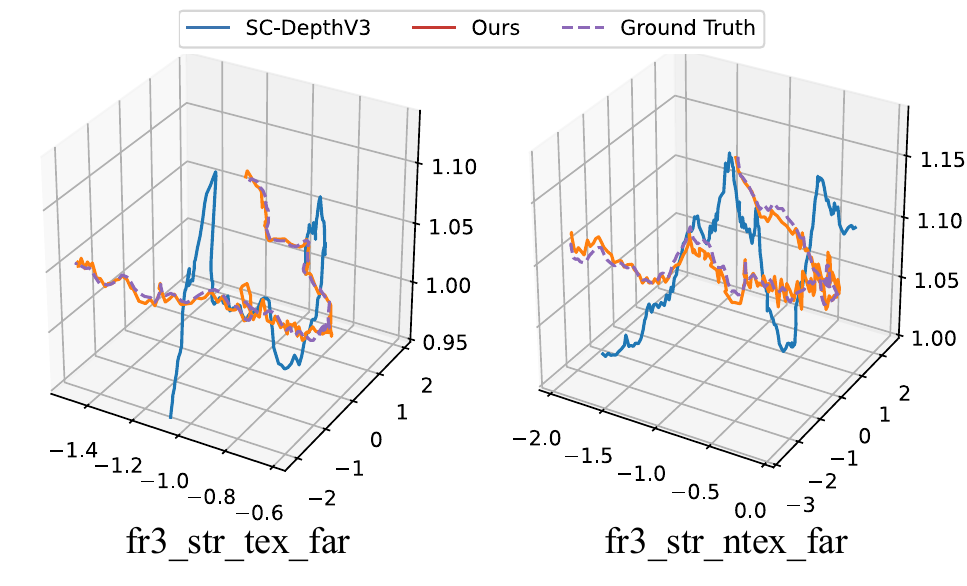}
      \caption{Qualitative results on the TUM RGB-D dataset.}
      \label{Fig_7}
\end{figure}

As shown in Table \ref{Tab_4}, supervised methods still deliver higher accuracy overall, and classical geometry-based SLAM pipelines frequently fail in dynamic or textureless sequences. Nonetheless, our self-supervised method till surpasses them on nearly every sequence. 
The performance gap originates from our Gumbel Optimization, directly addressing motion dynamics versus neural methods' implicit representations that falter under abrupt motions and occlusions. Trajectory visualizations in Figure \ref{Fig_7} quantitatively reinforce these advantages.

\begin{table}
\centering
\caption{\textbf{Odometry results (ATE) on the Euroc-Mav Dataset.} 
Methods are grouped into geometry‑based VO (Geo. M), geometry‑based VIO (Geo. M+I), and end-to-end VO (E2E VO).
`{\color{red}{$\times$}}' means that method fails in that sequence.
Best results are highlighted in \colorbox{Min1}{\textbf{first}}, \colorbox{Min2}{\underline{second}}, and \colorbox{Min3}{\textit{third}}, respectively.}
\begin{tabular}{clcccc}
\hline
                          \multicolumn{2}{c|}{Methods}                              & MH03  & MH05  & V103 & V202 \\ \hline
\multirow{2}{*}{Geo. M}   & \multicolumn{1}{c|}{DSO \cite{engel2017direct}}       & 0.18 & \cellcolor{Min3}\textit{0.11} & 1.42 & 0.12 \\
                          & \multicolumn{1}{c|}{ORB-SLAM2 \cite{mur2017orb}}       & \cellcolor{Min1}\textbf{0.08} & 0.16 & 1.48 & 1.72 \\ \hline
\multirow{5}{*}{Geo. M+I} & \multicolumn{1}{c|}{VINS \cite{qin2018vins}}      & 0.13 & 0.35 & \cellcolor{Min3}\textit{0.13} & \cellcolor{Min3}\textit{0.08} \\
                          & \multicolumn{1}{c|}{OKVIS \cite{leutenegger2015keyframe}}     & 0.24 & 0.47 & 0.24 & 0.16 \\
                          & \multicolumn{1}{c|}{ROVIO \cite{bloesch2015robust}}     & 0.25 & 0.52 & 0.14 & 0.14 \\
                          & \multicolumn{1}{c|}{MSCKF \cite{mourikis2007multi}}     & 0.23 & 0.48 & 0.24 & 0.16 \\
                          & \multicolumn{1}{c|}{SVO \cite{forster2016manifold}}       & 0.12 & 0.16 & \textcolor{red}{$\times$}    & \textcolor{red}{$\times$}    \\
\hline
\multirow{4}{*}{E2E VO}   & \multicolumn{1}{c|}{CC \cite{kuznietsov2021comoda}}        & 0.13 & 0.35 & 0.21 & 0.08 \\
                          & \multicolumn{1}{c|}{D3VO-Pose \cite{yang2020d3vo}} & 1.80 & 0.88 & 1.00 & 1.24 \\
                          & \multicolumn{1}{c|}{D3VO-Unce \cite{yang2020d3vo}} & \cellcolor{Min1}\textbf{0.08} & \cellcolor{Min2}\underline{0.09} & \cellcolor{Min2}\underline{0.11} & \cellcolor{Min2}\underline{0.05} \\
                          & \multicolumn{1}{c|}{Ours}      & \cellcolor{Min3}\textit{0.11} & \cellcolor{Min1}\textbf{0.08} & \cellcolor{Min1}\textbf{0.08} & \cellcolor{Min1}\textbf{0.04} \\ \hline
\end{tabular}
\label{Tab_3}
\end{table}

\begin{table*}[htbp]
\caption{\textbf{Odometry results (ATE) on the TUM-RGBD Odometry Dataset.} 
Methods are grouped into geometry‑based, supervised, and self‑supervised.
$\dagger$ is the official result of another training split, and  $\ddagger$ is our retrained version.
`{\color{red}{$\times$}}' means that method fails in that sequence.
Best results are highlighted in \colorbox{Min1}{\textbf{first}}, \colorbox{Min2}{\underline{second}}, and \colorbox{Min3}{\textit{third}}, respectively.}
\label{Tab_4}
\centering
\resizebox{\linewidth}{!}{
\begin{tabular}{l|cc|ccc|ccccc}
\hline
\multirow{2}{*}{Model}     & \multicolumn{2}{c|}{Geometry-based methods} & \multicolumn{3}{c|}{Supervised methods}                   & \multicolumn{5}{|c}{Self-Supervised methods}                          \\
                           & ORB‐SLAM2 \cite{mur2017orb}    & DSO \cite{engel2017direct}     & DeepV2D \cite{2020DeepV2D} & DeepV2D(aligned) \cite{2020DeepV2D} & Beyondtracking \cite{xue2019beyond} & LTMVO \cite{zou2020learning} & SC‐Depthv3 $\dagger$ & SC‐Depthv3 $\ddagger$ \cite{sun2023sc} & FrozenRecon \cite{sun2023sc} & Ours  \\ \hline
fr2/desk                   & \cellcolor{Min2}\underline{0.041}         &\textcolor{red}{$\times$}       & 0.232   & \cellcolor{Min3}\textit{0.087}            & 0.153          & 0.192 & 1.284      & 1.439       & 0.996       & \cellcolor{Min1}\textbf{0.026} \\
fr2/360\_kidnap            & \cellcolor{Min2}\underline{0.184}         & 0.197    & 0.651   & 0.300            & 0.208          & \cellcolor{Min3}\textit{0.190} & 1.018      & 1.161       & 1.449       & \cellcolor{Min2}\underline{0.188} \\
fr2/pioneer\_360           &\textcolor{red}{$\times$}            &\textcolor{red}{$\times$}       & 0.186   & 0.114            & \cellcolor{Min1}\textbf{0.056}          & \cellcolor{Min2}\underline{0.083} & 1.197      & 1.528       & 1.692       & \cellcolor{Min3}\textit{0.162} \\
fr2/pioneer\_slam3         &\textcolor{red}{$\times$}            & {0.737}    & 0.167   & \cellcolor{Min2}\underline{0.106}            & \cellcolor{Min1}\textbf{0.070}          & 0.122 & 1.794      & 1.879       & 2.068       & \cellcolor{Min2}\underline{0.120} \\
fr3/large\_cabinet         &\textcolor{red}{$\times$}            &\textcolor{red}{$\times$}       & \cellcolor{Min2}\underline{0.171}   & 0.181            & \cellcolor{Min3}\textit{0.172}          & 0.177 & 1.138      & 1.070       & 1.101       & \cellcolor{Min1}\textbf{0.074} \\
fr3/sitting\_static        &\textcolor{red}{$\times$}            & 0.082    & 0.029   & \cellcolor{Min2}\underline{0.013}            & 0.015          & 0.016 & \cellcolor{Min3}\textit{0.014}      & 0.016       & 0.023       & \cellcolor{Min1}\textbf{0.011} \\
fr3/nstr\_ntex\_near &\textcolor{red}{$\times$}            &\textcolor{red}{$\times$}       & 0.435   & 0.380            & \cellcolor{Min2}\underline{0.123}          & \cellcolor{Min3}\textit{0.219} & 0.846      & 1.297       & 1.412       & \cellcolor{Min1}\textbf{0.063} \\
fr3/nstr\_tex\_near  & \cellcolor{Min3}\textit{0.057}         & 0.093    & 0.106   & 0.110            & \cellcolor{Min1}\textbf{0.007}          & 0.102 & 1.718      & 1.679       & 0.754       & \cellcolor{Min2}\underline{0.055} \\
fr3/str\_ntex\_far         &\textcolor{red}{$\times$}            & 0.543    & \cellcolor{Min3}\textit{0.085}   & 0.094            & \cellcolor{Min2}\underline{0.035}          & 0.179 & 0.569      & 0.257       & 0.593       & \cellcolor{Min1}\textbf{0.016} \\
fr3/str\_tex\_far          & \cellcolor{Min3}\textit{0.018}         & \cellcolor{Min2}\underline{0.040}    & 0.082   & 0.098            & 0.042          & 0.107 & 0.388      & 0.454       & 0.296       & \cellcolor{Min1}\textbf{0.015} \\ \hline
\end{tabular}
}
\end{table*}

\begin{figure*}
      \centering
        \includegraphics[width=7in]{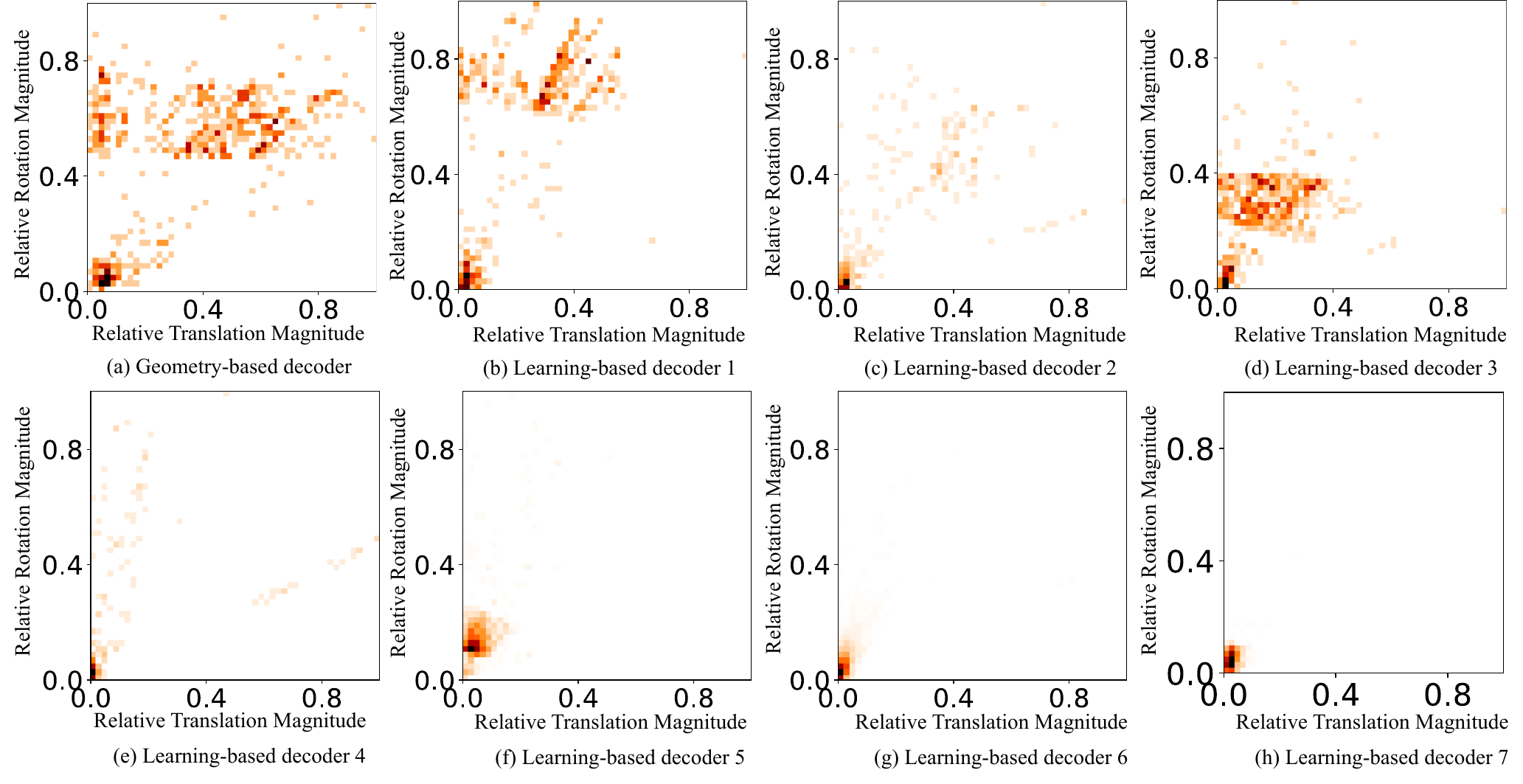}
      \caption{\textbf{Decoder activation for local state estimation.} We visualize the activation responses of eight specialized decoders on the KITTI dataset: one geometry-based decoder and seven learning-based decoders. Each panel shows how individual decoders are triggered by different translation and rotation magnitude.}
      \label{Fig_8}
\end{figure*}

\begin{table}[htbp]
\caption{\textbf{Ablation experiment of our methods.} The values describe the ATE.}
\centering
\begin{tabular}{c|cc|cc}
\hline
Methods     & MoE Decoder               & Gumb Opt.                 & KITTI & TUM-RGBD \\ \hline
Sc-Sfmlearner \cite{bian2019unsupervised} & $\times$                  & $\times$                  & 18.13      & -         \\
Sc-DepthV3 \cite{sun2023sc} & $\times$                  & $\times$                  &  -     & 0.72         \\\hline
Ours        & \checkmark & $\times$                  & 2.18      & 0.55         \\
Ours        & $\times$                  & \checkmark & 7.45      & 0.45         \\
Ours        & \checkmark & \checkmark &  1.58     & 0.04         \\ \hline
\end{tabular}
\label{Tab_6}
\end{table}

\begin{table}[]
\caption{Ablation experiments on the effect of varying the number of learning-based decoders.}
\centering
\begin{tabular}{c|cc|cc}
\hline
\multirow{2}{*}{Number of decoders} & \multicolumn{2}{c|}{Seq. 09} & \multicolumn{2}{c}{Seq. 10} \\ \cline{2-5} 
                   & $T_{rel}$     & $R_{err}$    & $T_{rel}$    & $R_{err}$ \\ \hline
1                & 7.32          & 3.05         & 7.79         & 4.90         \\
3                  & 5.21          & 2.21         & 4.21         & 2.81         \\
5                  & \cellcolor{Min1}\textbf{2.23}          & 1.42         & \cellcolor{Min1}\textbf{2.88}         & \cellcolor{Min1}\textbf{2.03}         \\
7                  & 2.34          & \cellcolor{Min1}\textbf{1.38}         & 2.96         & 2.10         \\ \hline
\end{tabular}
\label{Tab_7}
\end{table}

\subsection{Ablation study and visualization}
In this section, we conduct three sets of ablation and visualization experiments to examine the contributions of the different components of our framework. First, we assess the impact of each of our MoE-Decoder and Gumbel-based optimization modules on VO performance. Specifically, we compare our approach with two similar baselines: Sc-Sfmlearner \cite{bian2019unsupervised} on KITTI and SC-DepthV3 \cite{sun2023sc} on TUM-RGBD.
As shown in Table \ref{Tab_6}, on KITTI, which represents a relatively simple autonomous driving scenario, relying solely on the MoE decoder already achieves excellent performance.
However, on datasets representing more challenging motion dynamics, such as TUM-RGBD, while the MoE decoder provides some performance improvement, incorporating the Gumbel-based optimization module further refines the pose estimates.
These results demonstrate that our Gumbel-based optimization module with robust inter-frame constraints is crucial for achieving accurate pose estimation in complex motion scenes.

Second, we examine the effect of varying the number of learning-based decoders on performance using the KITTI dataset. We experiment with different configurations by adjusting the number of specialized decoders in our multi-expert framework. As shown in Table \ref{Tab_7}, our findings indicate that increasing the number of decoders improves pose estimation accuracy up to a certain point, beyond which the performance either plateaus or slightly decreases. This suggests that while multiple decoders are essential for capturing diverse motion patterns, an excessive number may introduce redundancy and complicate the optimization process.  

Finally, to provide an intuitive understanding of our method’s inter-frame processing, we visualize the frame-to-frame transformations handled by each decoder on the KITTI dataset (see Figure \ref{Fig_8}). This visualization demonstrates the model’s ability to dynamically assign specialized decoders based on distinct motion patterns, thereby enhancing its adaptability to varying scene dynamics. The results reveal several key insights.
First, certain decoders excel at capturing smooth, linear trajectories, while others better accommodate sharp turns and irregular movements.
Second, decoders relying on geometric processing typically tend to handle larger inter-frame motions, whereas the learning-based decoders are more adept at managing more challenging motions (e.g. pure rotational movements) that are difficult to capture. This observation contrasts with the performance reported in DF-VO \cite{zhan2021df}, where the effectiveness of such depth decoders was less clearly demonstrated.
Third, although almost are activation on nuanced motion, none of these individual predictions achieves full accuracy on its own.
This finding underscores the importance of our differentiable inter‐frame association and joint optimization module, which aggregates these fine‐grained but imprecise estimates into a cohesive, highly reliable trajectory.

\section{Conclusions}\label{sec7}

In this work, we presented UNO, a novel self-supervised monocular VO framework that robustly estimates camera trajectories across diverse environments and sensor platforms. Our method rethinks the conventional VO pipeline by adopting a decoder-centric module that employs a Mixture-of-Experts design to dynamically select specialized decoders. This is coupled with a differentiable inter-frame association mechanism based on the Gumbel-Softmax sampling trick, which facilitates the joint selection of optimal frame relationships and decoders. Additionally, we integrate a lightweight bundle adjustment module that refines pseudo-depth predictions from a pre-trained depth foundation model by jointly optimizing scale and shift parameters, thereby overcoming the limitations of affine-invariant depth outputs.

Extensive experiments on benchmark datasets such as KITTI, EuRoC-MAV, and TUM-RGBD demonstrate that our method significantly outperforms state-of-the-art methods. Our approach delivers superior accuracy in both translational and rotational pose estimation, particularly in challenging scenarios characterized by complex motion dynamics and dynamic environmental interference. Moreover, our method exhibits strong generalization capabilities across diverse data domains, paving the way for more versatile and reliable applications in autonomous navigation and robotics.

While our method demonstrates strong generalization across diverse motion paradigms and datasets, its performance remains inherently limited by the quality and diversity of the training data, making zero-shot deployment challenging. In future work, we aim to extend our framework to virtual environments, leveraging large-scale synthetic datasets with richer and more diverse motion patterns. By enabling virtual-to-real online adaptation, we hope to realize a universal monocular VO system capable of handling arbitrary real-world scenes without task-specific fine-tuning.


\bmsection*{Financial disclosure}

None reported.

\bmsection*{Conflict of interest}

The authors declare no potential conflict of interests.

\bmsection*{DATA AVAILABILITY STATEMENT}

The data and code that support the findings of this study are available on request from the corresponding author. The data and code are not publicly available due to privacy or ethical restrictions.

\bibliography{wileyNJD-Chicago}

\end{document}